# Capsule Network Performance With Autonomous Navigation

Thomas Molnar and Eugenio Culurciello

ECE, Purdue University, 610 Purdue Mall, West Lafayette, 47907, IN, USA

## Abstract

*Capsule Networks (CapsNets) have been proposed as an alternative to Convolutional Neural Networks (CNNs). This paper showcases how CapsNets are more capable than CNNs for autonomous agent exploration of realistic scenarios. In real world navigation, rewards external to agents may be rare. In turn, reinforcement learning algorithms can struggle to form meaningful policy functions. This paper's approach Capsules Exploration Module (Caps-EM) pairs a CapsNets architecture with an Advantage Actor Critic algorithm. Other approaches for navigating sparse environments require intrinsic reward generators, such as the Intrinsic Curiosity Module (ICM) and Augmented Curiosity Modules (ACM). Caps-EM uses a more compact architecture without need for intrinsic rewards. Tested using ViZDoom, the Caps-EM uses 44% and 83% fewer trainable network parameters than the ICM and Depth-Augmented Curiosity Module (D-ACM), respectively, for 1141% and 437% average time improvement over the ICM and D-ACM, respectively, for converging to a policy function across "My Way Home" scenarios.*

## 1. Introduction

Capsule Networks (CapsNets) were first presented by [1] to address shortcomings of Convolutional Neural Networks (CNNs). The max pooling operation used with CNNs reduces the spatial size of data flowing through a network and thus loses information. This leads to the problem that "[i]nternal data representation of a convolutional neural network does not take into account important spatial hierarchies between simple and complex objects" [1]. CapsNets resolve this by encoding the probability of detection of a feature as the length of their output vector. The state of the detected feature is encoded as the direction where that vector points. If detected features move around an image, then the probability, or vector length, remains constant while the vector orientation changes. The idea of a capsule resembles the design of an artificial neuron but extends it to the vector form to enable more powerful representational capabilities. A capsule may receive vectors from lower level capsules as an input and then performs four operations on the input: matrix multiplication of input vectors, scalar weighting of input vectors, sum of weighted input vectors and lastly a vector-to-vector nonlinearity. These operations are illustrated in Figure 1.

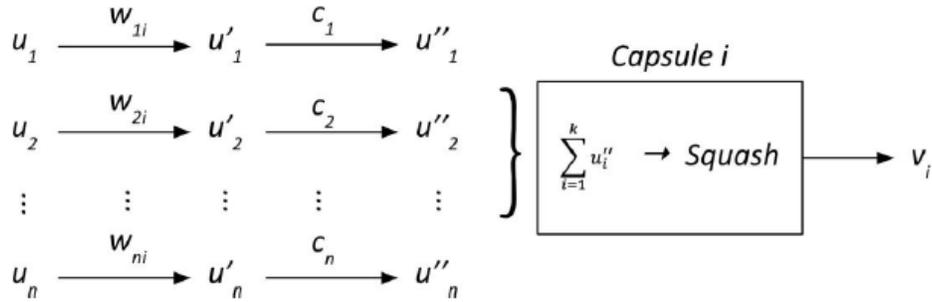

Figure 1. Illustration of Capsule Operations. A capsule receives input vectors $u_1$ through $u_n$. Vector lengths encode probabilities that lower-level capsules detected an object, while the vectors' directions encode the state of detected objects. An affine transformation with weight matrices $W_{1i}$ through $W_{ni}$ is applied to each vector. The weight matrices encode spatial and other relationships between lower level features and higher ones. After multiplication, the vectors $u'_1$ through $u'_n$ represent the predicted position of higher-level features. These vectors are multiplied by scalar weights $c_1$ to $c_n$, derived using the routing algorithm, to determine which higher-level capsule a lower level capsule's output maps to. The $k$ weighted input vectors $u''_1$ through $u''_n$ that map to Capsule $i$ are then summed to form one vector. The Squash nonlinear activation function takes the vector, forces it to length of max one, while not changing its direction, and outputs vector $v_i$ [1].

An approach called dynamic routing is the iterative method used to send lower-level capsule outputs to higher level capsules with similar outputs. The algorithm outlines how to calculate a network forward pass with capsules, as discussed by [1]. The method determines the vector $c_i$, which is all the routing weights for a given lower level capsule $i$. This is done for all lower level capsules. After this, the routing algorithm looks at each higher-level capsule, such as capsule $j$, to check each input and update weights in the formula. A lower level capsule tries to map its output to the higher-level capsule whose output is most similar. A dot product gauges the similarity between a capsule input and output. The algorithm repeats the process of matching lower level capsule outputs to the appropriate higher-level capsule $r$ times, where $r$ is the number of routing iterations.

Traditionally, reinforcement learning-based approaches for advancing autonomous agent navigation in realistic environments struggle to learn meaningful behavior. Reinforcement learning methods, such as Advantage Actor Critic (A2C), strive to maximize the amount of rewards that it obtains in an environment by learning an effective policy function. The rewards may vary given the environment and desired goal. With deep reinforcement learning, neural networks map input states to actions and are used to approximate a policy function. In environments with plentiful rewards for an actor to interact with, a neural network can readily update its policy function and converge to an optimal function for governing behavior. However in instances where rewards may be lacking and sparse, a network is not able to easily update its policy. In realistic scenarios that often have sparse rewards, a network can struggle to learn meaningful behavior and desired skills. This paper's approach called the Capsules Exploration Module (Caps-EM) is compared to previous research in addressing reinforcement learning shortcomings with exploring and navigating real world-like environments given sparse external rewards. [2] use the Intrinsic Curiosity Module (ICM) in union with an Asynchronous Advantage Actor Critic (A3C) algorithm to provide a reward signal intrinsic to an actor. This intrinsic reward supplements extrinsic reward signals that an actor encounters within an environment. When there are sparse external rewards, the intrinsic reward factor from the ICM still provides an agent with rewards independent of external rewards in an environment to still stimulate learning of new behavior and policy function updates. [3] similarly leverage intrinsic rewards with prediction of depth images in their Depth-Augmented Curiosity Module (D-ACM) to advance autonomous performance in sparse reward environments.

This paper demonstrates how CapsNets perform well for approximating policy functions when paired with an A2C algorithm to significantly improve autonomous agent navigation performance in sparse reward environments. Across a variety of test scenarios, the proposed Caps-EM uses a small network size to improve upon the ICM and D-ACM performances, which are presented as performance baselines. Critically relevant is the fact that Caps-EM does not incorporate the use of intrinsic rewards, which the ICM and D-ACM approaches both use to converge to adequate policy functions. This research highlights how strictly using external reward factors, Caps-EM achieves a more encompassing comprehension of image inputs and abstract world representation to achieve more meaningful action in any given scenario, which CNNs fail to replicate. While the Caps-EM struggles in certain test environments modeling extremely sparse external rewards, the module generalizes well across various scenarios with use of curriculum training and shows the capabilities of CapsNets in instances of real-world scenarios. Using a self-supervised framework, CapsNets advances autonomous system capabilities for navigation and exploration in challenging environments that can potentially be applied to robotics and Unmanned Aerial Vehicles (UAVs) for example.

## 2. RELATED WORK

Given that CapsNets are a recent development, published research on their applications is limited. [4] integrates CapsNets with Deep-Q Learning algorithms to evaluate performance across several different environments, including one with a task of exploring a maze. However, discussion of the architecture used by the author is limited, and the results show that CapsNets underperform traditional CNNs. Other work by [5] applies CapsNets to recurrent networks, and [6, 7, 8] successfully use CapsNets for image and object classification tasks. CapsNets have also been used for problems with autonomous driving in [9] and are effective with predicting depth for simultaneous localization and mapping (SLAM) implementations in [10]. [11] demonstrate how CapsNets may result in reduced neural network model training time and offer a lower number of training parameters relative to similar CNN architectures. [12] additionally highlight how capsules present more explainable internal operations than CNNs for better understanding of deep learning models. This paper's Caps-EM presents novel work on pairing CapsNets with an A2C algorithm specifically for autonomous exploration of and navigation through environments.

[2] propose the ICM to provide a supplemental intrinsic reward factor to an agent to handle sparse rewards in an environment. To generate this intrinsic reward, the ICM forms a prediction of the next state and compares the prediction to the ground state value of the next state. As shown in Figure 2, the ICM receives an action $a_t$, the current state $s_t$ and the next state $s_{t+1}$ as inputs. $a_t$ is the action taken by the agent to transition from $s_t$ to $s_{t+1}$. The current state and next state are RGB frames of the actor's view in VizDoom, a Doom-based platform for AI research. The ICM uses a forward model and an inverse model to generate the intrinsic reward factor. The forward model receives $a_t$ and attempts to predict an embedding of the next state $\varphi'_{t+1}$. The error between this

prediction and the ground truth label $\varphi_{t+1}$ obtained from $s_{t+1}$ is used as the intrinsic reward factor. The intrinsic reward is large when an agent explores new, unseen areas, as the predicted embedding is based on previously seen areas. This in turn rewards an agent to seek out new areas of environment. To train the forward model, the inverse model learns to predict the action $a'_t$ that relates the two state embeddings $\varphi_t$ and $\varphi_{t+1}$. [3] adapt the ICM and present the D-ACM that predicts depth images instead of an action in the inverse model. They contend that predicting depth images helps better encode 3D structural information in the embeddings. Their D-ACM outperforms the ICM with navigation tasks and is subsequently used an additional benchmark for comparison against the Caps-EM.

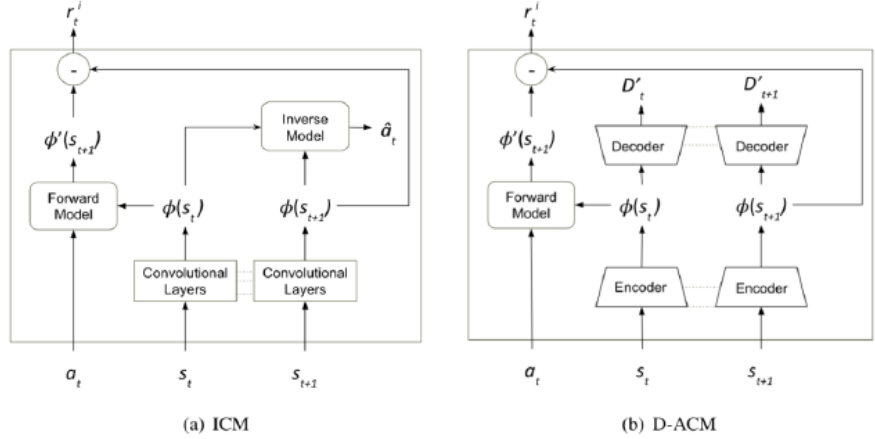

Figure 2. The ICM and D-ACM Architectures. (a) The ICM by [2] receives as inputs an action $a_t$ and the inputs states $s_t$ and $s_{t+1}$, which are 42x42 RGB images. Embeddings $\theta_t$ and $\theta_{t+1}$ are derived with a Convolutional Layer block. A Convolutional Layer block consists of four sets of 3x3 convolutions followed by batch normalization and an exponential linear unit (ELU) [13]. The Inverse Model uses the embeddings to predict action $a'_t$. Through the Forward Model, $\theta_t$ and label $a'_t$ are used to predict $\theta'_{t+1}$. The intrinsic reward factor r it is calculated as the difference between $\theta'_{t+1}$. and $\theta_{t+1}$. (b) The D-ACM by [3] receives the same inputs as the ICM but predicts the depth images $D'_t$ and $D'_{t+1}$ from each frame instead of an action. The Encoder and Decoder blocks have four layers of 32 filters with 3x3 kernels then equal numbers of filters and kernel sized convolutions. Dashed lines represent shared weights between networks.

## 3. METHODOLOGY

This next section discusses the Caps-EM approach, its network architecture design and how it operates with the A2C algorithm. Furthermore, the experiments and evaluation scenarios used to compare the various implementations are explained here.

### 3.1. Advantage Actor Critic (A2C) Algorithm

In reinforcement learning, a neural network controls an agent and strives to attain a maximal score by interacting with external reward factors in an environment. While [2] utilize an A3C algorithm with their ICM, an A2C algorithm is used by [3] and in this paper. The intrinsic reward factor generated by the ICM supplements the reward factor an agent receives from interactions with objects in its environment. The A2C paradigm consists of a critic, measuring the quality of an action, and an actor, which controls the agent's behavior. As the actor takes actions, the critic observes these actions and provides feedback. From this feedback, the actor updates its policy to improve performance.

At each time-step $t$, the current state $s_t$ of the environment is given as an input to the actor and critic. The actor, governed by the policy function $(s_t, a_t, \theta)$, with state $s_t$, action $a_t$ and network parameters $\theta$, receives $s_t$ and outputs the action $a_t$. The policy then receives the next state $s_{t+1}$ as well as the reward $r_{t+1}$ after the action is taken. In the evaluation environments, the actor is limited to a discrete action space consisting of four possible actions: move forward, move left, move right and no action. The critic, expressed as the value function $\hat{q}(s_t, a_t, w)$ with parameters $w$, returns an estimate of the expected final total reward obtainable by the agent from the given state. The value function $V_v(s_t)$, with network parameters $v$, returns the average value of a given state. A2C methods offer a variant for the value estimate to reduce the problem of high variability.

The advantage function, as shown in Equation 1, with $\gamma$ the discount factor to account for future rewards losing value, indicates the improvement of a taken action over the average action taken at the state.

$$A\ (s_t, a_t) = r_{t+1} + \gamma V_v\ (s_{t+1}) - V_v\ (s_t) \tag{1}$$

$$\Delta\theta = \alpha \nabla_\theta\ (log\pi_\theta(s_t, a_t))\hat{q_w}(s_t, a_t) \tag{2}$$

$$\Delta w = \beta(r(s_t, a_t) + \gamma \hat{q_w}(s_{t+1}, a_{t+1}) - \hat{q_w}(s_t, a_t))\nabla_w \hat{q_w}(s_t, a_t) \tag{3}$$

The actor and critic exist as separate models that are trained and optimized individually. The policy update for the network parameters, or weights $\theta$, uses the q value of the critic as shown in Equation 2. The critic updates its value parameters using the actor's output action $a_{t+1}$ for the next state $s_{t+1}$ in Equation 3. The hyperparameter $\beta$ controls the strength of entropy regularization.

A2C and A3C algorithms operate with the premise described above, however A3C algorithms asynchronously execute different agents in parallel on multiple instances of the environment. These agents update the globally shared network asynchronously [14]. With A2C the update occurs synchronously when all agents have completed training and uses the workers' averaged gradients to modify the global network at one time [15]. An A2C algorithm alone cannot converge to an optimal policy function in sparse scenarios, like those used in this paper for evaluation. Existing approaches with A2C require the use of intrinsic reward factors, as discussed in Figure 2. However, this paper's approach Caps-EM can use an A2C algorithm without supplementary intrinsic rewards to effectively explore extremely sparse scenarios.

## 3.2. Capsules Exploration Module (Caps-EM) Architecture

CapsNets initially appeared advantageous for the task of exploration even prior to testing. They can discern both the probability of detection of a feature, stored in an output vector's magnitude, and the state of the detected feature, stored in a vector's direction [1]. Conversely, traditional CNNs are only able to handle the probability of detection of a feature. This difference proves vital as CapsNets can then maintain spatial relationships of observed items in environment. This distinction hypothetically enables creation of more sophisticated network embeddings of the environment space. This paper demonstrates experimentation combining CapsNets with A2C components as not previously explored in other published literature. Importantly, Caps-EM does not use intrinsic rewards like the approaches discussed in Figure 2. The architecture implementation of the Caps-EM is illustrated in Figure 3. It is important to note as well that using an A2C network design exclusively incorporating CNNs, with no use of intrinsic reward signals, cannot explore effectively. In testing, such an approach failed to learn and converge to an effective policy function in any of the evaluation scenarios discussed in this paper.

As done by [1], the network proposed in Figure 3 uses a convolutional layer before the capsule layers to detect basic features in the 2D RGB image inputs. The subsequent capsule layer then uses the detected basic features to produce combinations of the features. From experimentation with various architecture designs, using more than one convolution before the capsule layers masks the benefits of the capsule layers by lowering the data resolution and degrading performance.

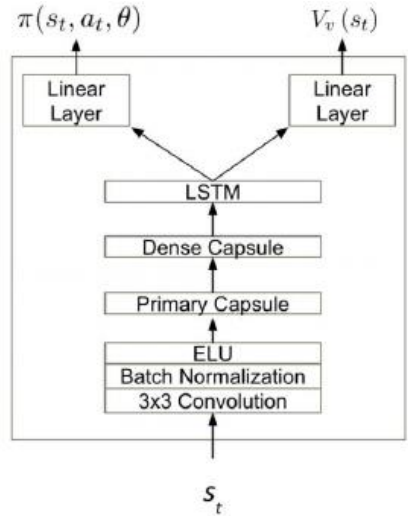

Figure 3. Caps-EM Advantage Actor Critic Network. In the Caps-EM, the input $s_t$, a 42x42 RGB image, first passes through a series of a 3x3 convolution, batch normalization function and ELU. The remaining layer blocks consist of Capsule Network layers. The first lower level Primary Capsule layer consists of a 9x9 convolution, with stride of two and padding of zero, followed by a Squash activation function [1]. This layer has 32 input and output channels with capsule dimension of eight. The outputs are dynamically routed to the second higher level Dense Capsule layer consisting of 196 input capsules of dimension eight and four output capsules of dimension 16. Three routing iterations are used in the routing algorithm. Outputs of the Dense Capsule layer are passed to an LSTM and linear layers to provide the policy function $\pi(s_t, a_t, \theta)$ and state value function $V_v(s_t)$.

Additionally, using a larger network architecture with more trainable network parameters was not found to increase the module's performance in converging to an optimal policy function or with being more generalizable. In fact, using a larger network architecture degrades overall performance due to the network model needing longer to train to converge to a policy. Three routing iterations are used between capsule layers as recommended by [1] to help prevent overfitting the data. However in experiments, the capsule layers still displayed a tendency for overfitting. In these instances, the early stopping method was used to avoid overfitting when the network successful achieved an adequate policy function [16]. Neither did incorporating dropout significantly improve the problem of network overfitting [17]. Dropout with p values of 0.25 and 0.5 were applied to various layers of the Caps-EM module, with the main effect only being a slowed training rate. The architecture described in Figure 3 for Caps-EM was found to be one that balanced the desire for a generalizable network across all evaluation scenarios with also a minimal number of network parameters.

## 3.3. Evaluation Environments

Similarly as with [3], ViZDoom is used as the evaluation environment to assess the modules' capabilities [18]. Within ViZDoom several scenarios are utilized to frame environments with sparse external rewards. Across the various scenarios, an agent is tasked to search for a piece of armor that serves as the sole positive external reward. A scenario restarts after an agent reaches its goal, the armor, and receives a reward factor of +1 or after surpassing 2100 time steps. The standard scenarios used are the "MyWayHome-v0" (MWH) setups from OpenAIGym [19]. MWH has 8 rooms with unique wall textures. Recreating the evaluation process of [3], two additional scenarios named "My Way Home Mirrored" (MWH-M) and "My Way Home Giant" (MWH-G) are used for testing the modules. MWH-M helps to judge how well a module's learned network knowledge transfers between scenarios for its generalizability. The scenario consists of the same number of rooms as MWH, however the layout is rotated. MWH-G is similar to MWH but with 19 rooms, thus presenting a much more complex and challenging evaluation environment. These different scenarios are shown in Figure 4.

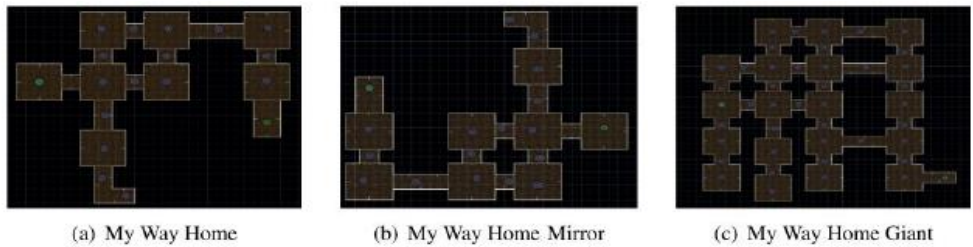

(a) My Way Home      (b) My Way Home Mirror      (c) My Way Home Giant

Figure 4. ViZDoom Scenarios. Green circles at the right in the images are the target locations. For dense settings, an agent may start at any purple circle. For sparse settings, an agent starts at the green circle at the left of each scenario [3].

Variability is further introduced to the evaluation process using variety in reward sparsity as well as in visual texture features [2]. To incorporate variations in scenario complexity, scenarios exhibit either a sparse or dense structure. This targets a given scenario's reward sparsity. With the target location remaining unchanged, the agent's beginning position is allowed to vary in the dense case; there are 17 uniformly distributed available start positions. However with sparse, the position is singular and placed far away from the goal location.

Regarding visual texture features of scenarios, the wall textures of an environment may either be constant and all identical or vary between each room. These two types of setups are noted as Uniform Texture and Varied Texture, respectively, further on. These variants are demonstrated in Figure 5. In the context of Uniform Texture, an embedding must maintain an abstract interpretation of the whole environment to effectively explore and find the target without receiving textural cues from the environment.

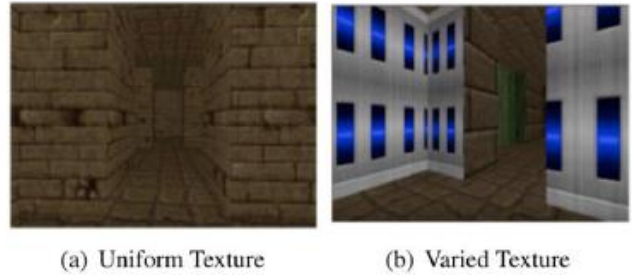

Figure 5. Variability in Visual Texture Features. (a) Shows a frame from the agent's point of view in a Uniform Texture scenario. (b) Agent's view of a Varied Texture scenario [3].

### 3.4. Curriculum Learning

To explore how well Caps-EM applies knowledge obtained in a successful policy function from one scenario to another, curriculum learning is used in junction with the MWH-M scenario [20]. For this procedure, a module is trained until converged to a policy in each of the MWH Dense Varied Texture, Dense Uniform Texture, Sparse Varied Texture and Sparse Uniform Texture scenarios. The learned network parameter values associated with these respective scenarios are then pre-loaded into the same module prior to beginning training again in the respective MWH-M Dense Varied Texture, Dense Uniform Texture, Sparse Varied Texture and Sparse Uniform Texture scenarios. In this way, the module begins training in the MWH-M environments with prior knowledge learned in the MWH scenarios. The various modules are also allowed to train in MWH-M without use of curriculum learning, and the results of the two different approaches are compared.

## 4. RESULTS

For analyzing the Caps-EM, the ICM by [2] and the D-ACM by [3] are used as baselines for comparison. Table 1 compares these three approaches, where the percent difference rows indicate a module's improved or degraded metrics relative to the ICM. The number of trainable parameters of each module are used as a comparison metric to account for the differences in module size and scaling. The size, or number of trainable parameters, of each module has a direct impact on the efficiency and required time to complete the neural network model training process. The Caps-EM architecture has substantially fewer trainable network parameters, in turn completing a single training step more efficiently as well. The times to complete one training step as shown are standardized values obtained from running each module variation with one worker on a GeForce GTX 1080 GPU with 8114 MiB memory on the MWH Dense Varied Textured scenario. Results tables displayed further on showing timing analysis are based on these standardized times to present an equivalent metric of comparison. Plots are presented with the mean testing score of a module relative to the number of training steps taken.

Table1. Model Size and Timing Comparisons.

| Module | # of Trainable Parameters | Time to Complete 1 Training Step (ms) |
|---|---|---|
| ICM | 915,945 | 13.6 |
| D-ACM | 944,170 | 17.6 |
| % Difference | +3% | +30% |
| Caps-EM | 515,301 | 6.2 |
| % Diff. | -44% | -54% |

As shown in Figure 6 and Table 2, the Caps-EM performs exceptionally well in the dense setup scenarios. While the Caps-EM completes the MWH Sparse Uniform scenario in roughly the same number of training steps as D-ACM, the Caps-EM performance is superior when considering the actual time to converge to a policy and how Caps-EM does not use intrinsic rewards. Conversely in the Sparse Varied Texture scenario, the Caps-EM performs worse than both the ICM and D-ACM.

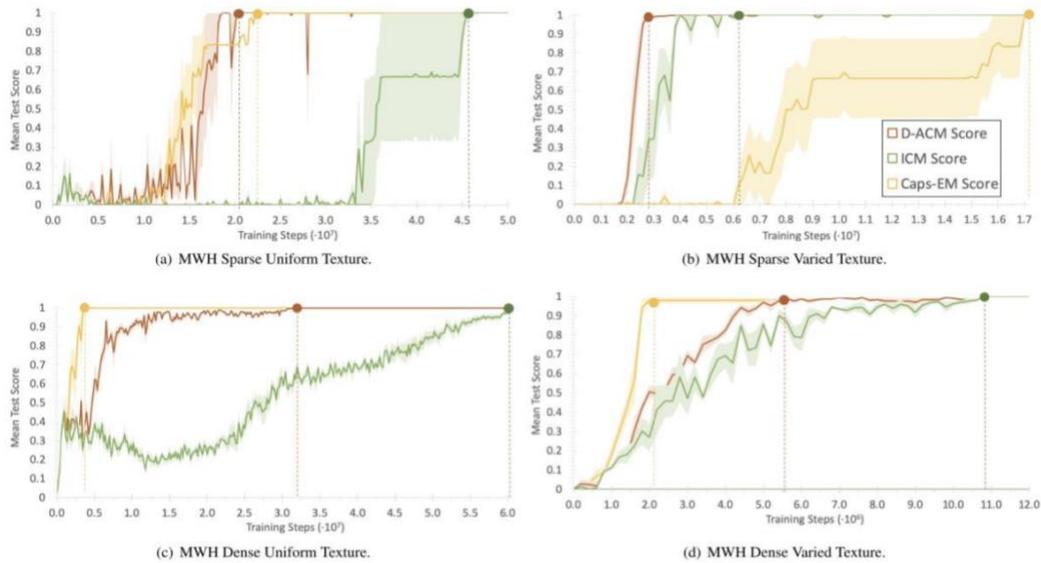

Figure 6. My Way Home Results. A minimum of five instances of a module's performance is averaged for the score trend line. Shaded areas around trend line indicates the one standard error range, and ovals roughly indicate where a module converges to a policy function.

Table 2. My Way Home Scenario Time Performance Results. Time required to converge to an optimal network policy is in seconds.

| Module | Scenario Dense, Varied | Dense, Uniform | Sparse, Varied | Sparse, Uniform |
|---|---|---|---|---|
| ICM (s) | 1.46E+5 | 8.15E+5 | 5.44E+4 | 6.18E+5 |
| D-ACM (s) | 9.69E+4 | 2.64E+5 | 4.85E+4 | 3.52E+5 |
| % Diff. | -51% | -209% | -12% | -75% |
| Caps-EM (s) | 1.24E+4 | 2.48E+4 | 1.07E+5 | 1.37E+5 |
| % Diff. | -1076% | -3182% | +49% | -353% |

In order to assess how integrated intrinsic rewards with Caps-EM affects the module's performance with exploration, Figure 7 shows a direct comparison of Caps-EM with and without intrinsic reward in the same scenarios. The Caps-EM with 515,301 trainable parameters is 42% smaller than Caps-EM with intrinsic rewards, referred to as Caps-EM (IR), which has 733,705 trainable parameters. Caps-EM (IR) incorporates the approach discussed in Figure 2a to generate an intrinsic reward based on the accuracy of next state predictions. Caps-EM requires 6.21E-3 seconds to complete one training step, whereas Caps-EM (IR) takes 2.09E-2 seconds and is 236% slower. Table 3 shows a comparison of performance with respect to time across the MWH scenarios and that Caps-EM (IR) exhibits poorer performance in each setup. In this analysis, intrinsic reward do not necessarily improve performance in the Sparse Varied Texture scenario.

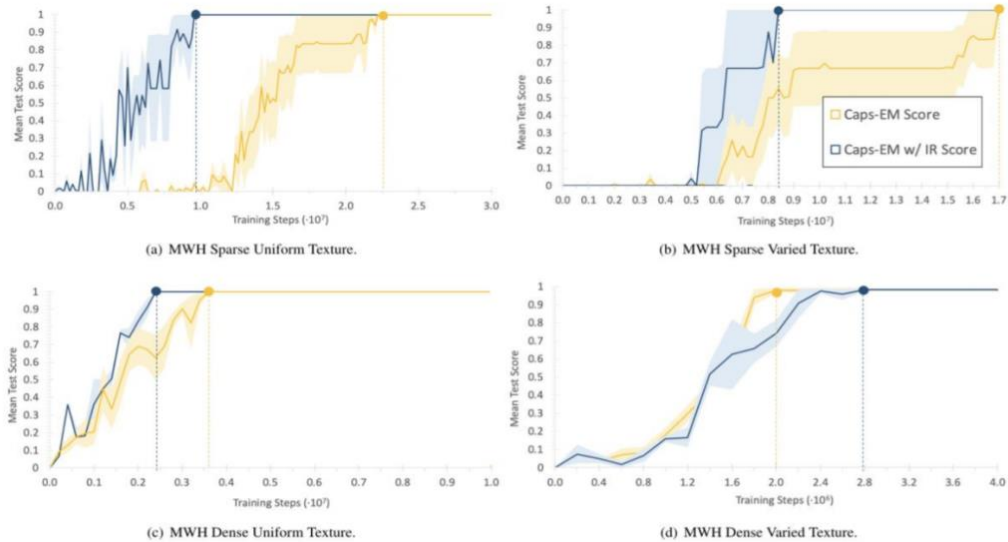

Figure 7. My Way Home Caps-EM, with and without Intrinsic Rewards (IR), Results.

Table 3. My Way Home Scenario Caps-EM, with and without IR, Time Performance Results.

| Module | Scenario | | | |
| --- | --- | --- | --- | --- |
| | Dense, Varied | Dense, Uniform | Sparse, Varied | Sparse, Uniform |
| Caps-EM (s) | 1.24E+4 | 2.48E+4 | 1.07E+5 | 1.37E+5 |
| Caps-EM (IR) (s) | 5.74E+4 | 5.22E+4 | 1.78E+5 | 2.09E+5 |
| % Diff. | +78% | +52% | +40% | +35% |

Figure 8 and Table 4 show how well each module applies learned network parameter weights from the MWH scenarios to the MWH-M scenarios. The expectation is that the knowledge should generalize well and enable the modules to converge to a successful policy faster than without using curriculum learning. The Caps-EM fails to converge to a policy function in the MWH-M Sparse Varied Texture and Sparse Uniform Texture scenarios within 1.0E+8 training steps when not using curriculum learning. This may due to the how extremely sparse these respective scenarios are in design, combined with how the Caps-EM lacking an intrinsic reward signal to motivate exploration. Experiments showed that Caps-EM (IR) was able to converge to a policy in roughly 2.5E+7 training steps in MWH-M Sparse Uniform Texture and in 3.0E+7 steps in MWH-M Sparse Varied Texture with no curriculum training. However when using curriculum learning in these same scenarios, the Caps-EM without intrinsic rewards performs exceptionally well. In general, each module leverages curriculum learning to its advantage and can outperform the no curriculum learning scores.

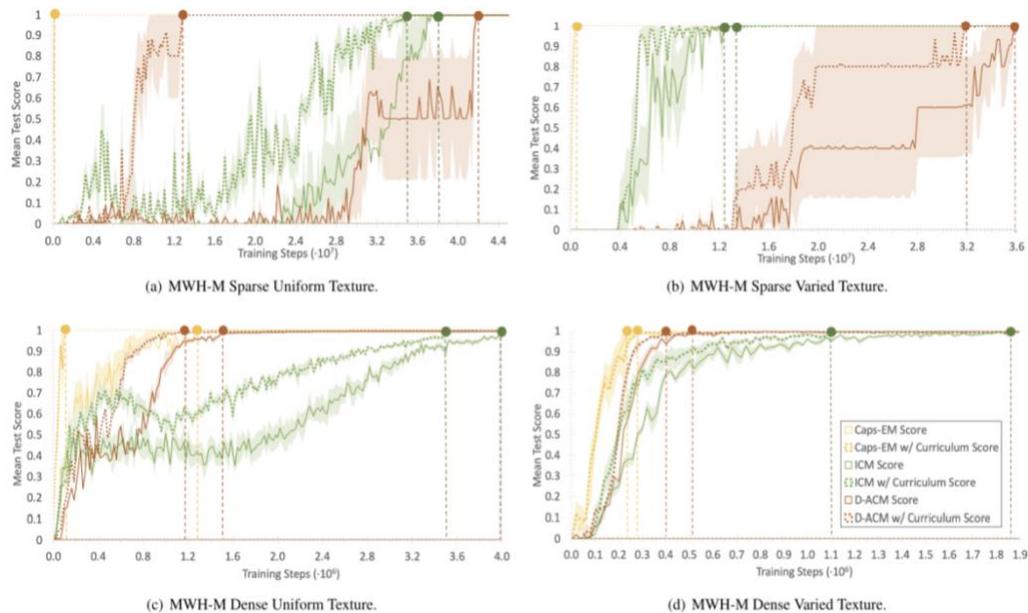

Figure 8. My Way Home Mirrored with Curriculum Learning Results.

Table 4. My Way Home Mirrored Scenario Time Performance Results, with and without Curriculum Learning. Positive percent difference values indicate the improved runtime from curriculum training.

| Module | Scenario | | | |
|---|---|---|---|---|
| | Dense, Varied | Dense, Uniform | Sparse, Varied | Sparse, Uniform |
| ICM (s) w/ Curriculum Training | 1.49E+5 | 4.76E+5 | 1.70E+5 | 4.76E+5 |
| w/o Curr. Training | 2.45E+5 | 5.44E+5 | 2.04E+5 | 5.10E+5 |
| % Diff. | +39% | +13% | +17% | +7% |
| D-ACM (s) w/ Curr. Training | 7.93E+4 | 1.94E+5 | 5.46E+5 | 2.20E+5 |
| w/o Curr. Training | 1.15E+5 | 2.64E+5 | 6.26E+5 | 7.49E+5 |
| % Diff. | +31% | +27% | +13% | +71% |
| Caps-EM (s) w/ Curr. Training | 1.71E+3 | 6.21E+3 | 2.48E+3 | 1.24E+3 |
| w/o Curr. Training | 1.86E+3 | 8.07E+3 | - | - |
| % Diff. | +8% | +23% | - | - |

Figure 9 and Table 5 demonstrate each module's performance in MWH-G. None of the modules converge to a successful policy function in the sparse scenarios variants, with or without use of curriculum learning. The Caps-EM for example reached in excess of 12.6E+7 training steps without any learning of a useful function. MWH-G results illustrate how the Caps-EM performs well in dense scenario variants and significantly outperforms the ICM and D-ACM which both depend on use of intrinsic rewards.

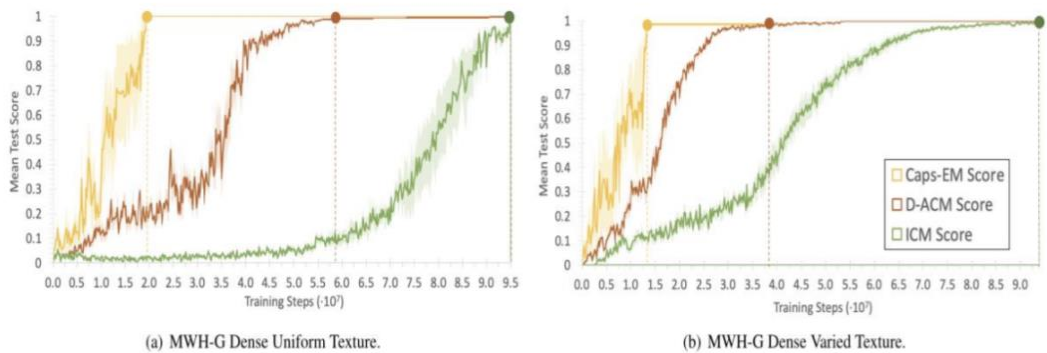

(a) MWH-G Dense Uniform Texture.     (b) MWH-G Dense Varied Texture.

Figure 9. My Way Home Giant Results.

Table 5. My Way Home Giant Scenario Time Performance Results.

| Module | Scenario | |
|---|---|---|
| | Dense, Varied | Dense, Uniform |
| ICM (s) | 1.26E+6 | 1.29E+6 |
| D-ACM (s) | 6.70E+5 | 9.16E+5 |
| % Diff. | -88% | -41% |
| Caps-EM (s) | 8.07E+4 | 1.18E+5 |
| % Diff. | -1457% | -994% |

## 5. CONCLUSIONS

The Caps-EM architecture leverages the A2C scheme to perform well with autonomous navigation and exploration in sparse reward environments. More compact and efficient with 44% and 83% fewer parameters than the ICM and D-ACM, respectively, the Caps-EM on average outperforms both the ICM and D-ACM across the MWH, MWH-M and MWH-G scenarios. The Caps-EM converges to a policy function in MWH, on average across all four scenario variants, 437% and 1141% quicker than the D-ACM and ICM, respectively, without the use of intrinsic rewards. Similarly in MWH-M scenarios when using curriculum learning, the Caps-EM has a 10,726% and 13,317% time improvement on average over the D-ACM and ICM, respectively. Lastly with MWH-G variants, the Caps-EM has a 703% and 1226% time improvement on average over the D-ACM and ICM, respectively.

While the Caps-EM struggles to converge effectively in certain sparse scenarios, such as with MWH-M Sparse Uniform Texture and Sparse Varied Texture, the module readily applies learned knowledge using curriculum learning to generalize well across scenarios. The intrinsic reward factor used by the D-ACM and ICM likely enables these modules to better handle these specific sparse scenarios. Yet, the Caps-EM (IR) module did not significantly improve performance in these scenarios. However, these modules that produce the intrinsic signal must be trained in addition to the A2C algorithm itself. In turn, these approaches have larger architectures with more network parameters and lower relative performance in other scenarios. Caps-EM offers a more lightweight yet still capable design.

The results additionally confirm the hypothesis of Caps-EM's ability to maintain better spatial relationships and hierarchies for improved performance on average. This finding is evident in dense scenarios where the Caps-EM maintains relationships between the Varied Texture rooms well. Future work will explore how to improve the Caps-EM performance in the extremely sparse environments, such as MWH-M Sparse Uniform Texture, to address this weakness. Experimentation with Caps-EM variants that did incorporate intrinsic reward factors did improve effectiveness in these edge cases to a degree. However, this module variant does not appear to be a viable solution as its performance on average in other scenarios is significantly worse. This finding arises from how the addition of intrinsic rewards requires a larger network and how any improvements in navigation only prove applicable in limited cases. An additional area of interest is to study how different inputs than only 2D RGB frames would affect the Caps-EM performance with exploration tasks. Moreover, the scenarios used for evaluation the module are static with no moving objects or features, which could also have an impact on performance and be useful to investigate.